\documentclass[runningheads]{llncs}

\usepackage{eccv}
\usepackage{eccvabbrv}

\usepackage{graphicx}
\usepackage{booktabs}

\usepackage[accsupp]{axessibility} 
\usepackage{multirow} 
\usepackage{indentfirst}
\usepackage{float}
\usepackage{tabularx}
\usepackage{array}   
\usepackage{subcaption}
\usepackage{tabularray}
\usepackage{dsfont}
\usepackage{pifont}
\usepackage[misc]{ifsym}
\newcommand{\envelope}{\ding{41}}

\usepackage[pagebackref,breaklinks,colorlinks,citecolor=eccvblue]{hyperref}
\usepackage{orcidlink}

\newcommand{\name}{OSP}
\newcolumntype{Y}{>{\centering\arraybackslash}X}

\begin{document}

\title{Occupancy as Set of Points} 

\author{Yiang Shi\inst{1,\star}\and
Tianheng Cheng\inst{1, \star} \and
Qian Zhang\inst{2} \and \\
Wenyu Liu\inst{1} \and
Xinggang Wang\inst{1,\text{\envelope}}
}

\authorrunning{Y. Shi, X. Wang et al.}
\institute{School of EIC, Huazhong University of Science \& Technology \quad \and
Horizon Robotics\\
}

\maketitle
\let\thefootnote\relax\footnotetext{$^\star$ Equal contribution.}
\footnotetext{$^\text{\envelope}$ Corresponding author: Xinggang Wang (\url{xgwang@hust.edu.cn}).}

\begin{abstract}
In this paper, we explore a novel point representation for 3D occupancy prediction from multi-view images, which is named \textit{Occupancy as Set of Points}.
Existing camera-based methods tend to exploit dense volume-based representation to predict the occupancy of the whole scene, making it hard to focus on the special areas or areas out of the perception range. In comparison, we present the \textit{Points of Interest} (PoIs) to represent the scene and propose OSP, a novel framework for point-based 3D occupancy prediction. Owing to the inherent flexibility of the point-based representation, OSP achieves strong performance compared with existing methods and excels in terms of training and inference adaptability. It extends beyond traditional perception boundaries and can be seamlessly integrated with volume-based methods to significantly enhance their effectiveness. Experiments on the Occ3D-nuScenes occupancy benchmark show that OSP has strong performance and flexibility. Code and models are available at \url{https://github.com/hustvl/osp}.

\keywords{3D Occupancy Prediction \and Autonomous Vehicles \and Multi-view 3D Perception}
\end{abstract}

\section{Introduction}
\label{sec:intro}
Holistic 3D scene understanding is crucial for autonomous driving systems, directly affecting the efficiency and accuracy of subsequent tasks. Considering the cost-effectiveness and ease of deployment of cameras compared to other sensors, developing visual-based methods for 3D scene understanding has become a significant and widely researched challenge. 

To tackle this challenge, 3D Semantic Scene Completion (SSC)~\cite{song2017semantic} has been proposed and widely studied to jointly infer the geometry and semantics information of the scene from limited observations. The SSC task requires the model to accurately predict the visible locations and complete the information for the invisible locations. Recently, Occ3D \cite{tian2023occ3d} introduces a new task definition called 3D occupancy prediction. The main difference between this task and SSC is that 3D occupancy prediction only focuses on the visible areas and is tailored for dynamic scenes.

\begin{figure}[t]
  \centering
   \includegraphics[width=0.9\textwidth]{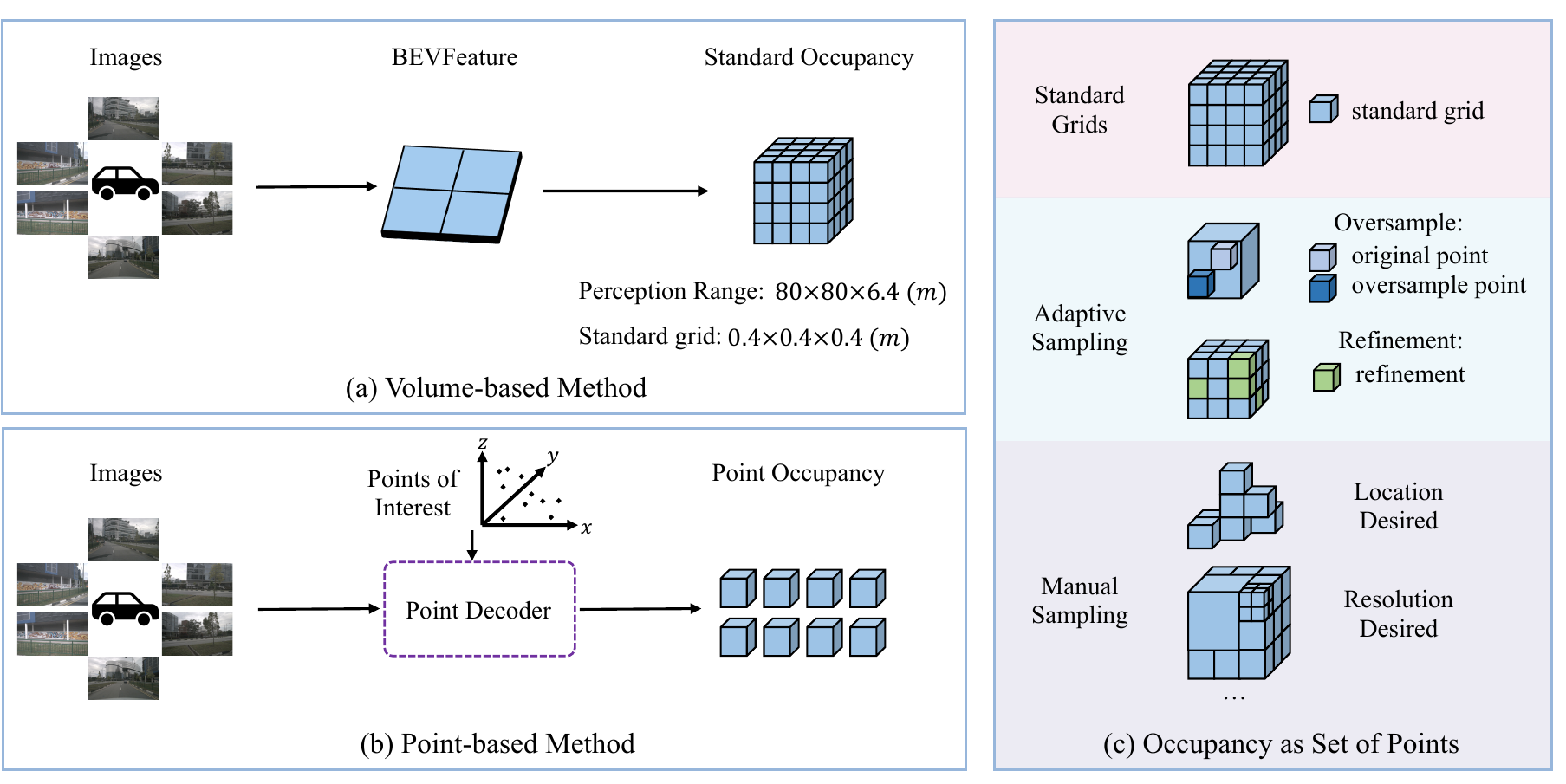}
   \caption{\textbf{Comparison between volume-based methods and our method.} The volume-based methods, represented by BEVFormer, infers every region within the scene and gets standard occupancy as shown in (a). Our method uses a point-based decoder as shown in (b). Thus it infers the \textbf{Points of Interest} including standard, adaptively sampled, and manually sampled grids as shown in (c).}
   \label{fig:teaser}
\end{figure}

Existing 3D occupancy prediction methods are mostly based on dense BEV methods, \eg, BEVFormer~\cite{li2022bevformer}, BEVDet \cite{bevdet}. These methods integrate a BEV encoder with an occupancy head to generate the output and enhance BEV perception capabilities for better results. However, they share some common drawbacks. (1) \textit{Uniform Sampling}: BEV-based methods fail to differentiate between different areas within the same scene, treating them equally. This leads to coarse sampling and hinders dynamic or multi-resolution sampling capabilities. (2) \textit{Limited Inference Flexibility}: During inference, these methods can only process the entire scene at once. They lack the ability to infer different parts of the scene based on varying downstream tasks or specific practical needs.

Those limitations highlight the need for more flexible 3D occupancy prediction methods that can handle complex scenes while adapting to different inference requirements. In this paper, we propose a novel point-based representation for 3D occupancy prediction. 
Instead of dividing the scene into uniform grids applied by existing volume-based methods, we propose \textit{Points of Interest} (PoIs) to view the scene as a collection of points that help in flexibly sampling the scene during both training and inference stages. 
Fig. \ref{fig:teaser} compares volume-based and point-based representations. Compared to volume-based representations, our point-based representation has the following advantages: (1) it can accept inputs of any scale and position to make occupancy predictions, including manually designed and adaptively designed input, offering flexibility; (2) it can pay extra attention to certain areas rather than treating all areas equally, enhancing the model's perceptual capabilities. 

We introduce Occupancy as Set of Points (\name{}), a novel and flexible point-based framework, which is built upon the foundational concept of Points of Interest (PoIs). \name{} excels in 3D occupancy prediction and is composed of an image backbone, a 3D positioning encoder, and a decoder, as illustrated in Fig. \ref{fig:main}. Central to our methodology is the innovative use of PoIs, which we have categorized into three distinct types to meet diverse needs, thereby significantly enhancing various aspects of our model's performance. Detailed descriptions of these PoIs will be provided in Sec. \ref{subsec:pre}. Notably, PoIs can be designed as needed beyond the three types initially proposed by us.

Our method stands out for its strong performance and flexibility. The flexibility enables it to process any arbitrary local scene without necessitating retraining. Additionally, our method can serve as an augmentative plugin module for existing volume-based methods by adaptively resampling areas of low confidence to yield more accurate occupancy predictions. It is also adept at predicting areas beyond the scene. The key contributions of our approach can be summarized as follows:
\begin{itemize}
    \item A novel point-based occupancy representation, established by interacting point queries with 2D image features, enables a comprehensive understanding of 3D scenes.
    \item A flexible framework that allows for inference at any area of interest without retraining or sacrificing accuracy and predicts areas beyond the scene.
    \item A plugin module that enhances the performance of the volume-based baseline significantly.
    \item \name{} has obtained strong experiment results, \ie, 39.4 mIoU in the 3D occupancy prediction task on the Occ3D-nuScenes benchmark. 
\end{itemize}

\section{Related Work}

\subsection{3D Occupancy Prediction}
3D occupancy prediction, a concept recently defined by Occ3D \cite{tian2023occ3d}, exhibits notable parallels with Occupancy Grid Mapping (OGM) \cite{moravec1985high,thrun2002probabilistic,wang2023openoccupancy} used in robotics. This task aims to predict the state of each voxel grid in a scene based on a series of sensor inputs. Occ3D establishes two benchmarks leveraging the Waymo Open Dataset \cite{sun2020scalability} and the nuScenes Dataset \cite{caesar2020nuscenes} to facilitate this. In vision-based 3D Occupancy prediction, Occ3D implements camera visibility estimation and creates visibility masks to ensure evaluations are confined to visible areas. It also evaluates various SSC methodologies on its benchmarks, including MonoScene \cite{cao2022monoscene}, TPVFormer \cite{huang2023triperspective}, BEVDet \cite{bevdet}, OccFormer \cite{zhang2023occformer}, and BEVFormer \cite{li2022bevformer}.

\subsection{3D Semantic Scene Completion}
Scene Semantic Completion (SSC) represents a task closely associated with 3D occupancy prediction. The concept of SSC is initially presented in SSCNet \cite{song2017semantic}, with a focus on predicting the comprehensive semantic information of a scene based on its partially visible regions. Over recent years, the study of SSC has expanded significantly, particularly in the context of small indoor scenes
\cite{zhang2018efficient,liu2018see,li2019depth,li2019rgbd,zhang2019cascaded,li2020anisotropic,chen20203d,cai2021semantic}. 
In recent times, the study of Scene Semantic Completion (SSC) for expansive outdoor environments has gained momentum, particularly following the introduction of the SemanticKITTI dataset \cite{behley2019semantickitti}. Notably, MonoScene \cite{cao2022monoscene} emerges as the first method to apply monocular pure vision-based SSC. In a parallel advancement, OccDepth \cite{miao2023occdepth} enhances 2D to 3D feature transformation by incorporating depth data from stereo input. TPVFormer \cite{huang2023triperspective} argues against the limitations of single-plane modeling in capturing intricate details, hence it adopts a tri-perspective view (TPV) approach, combining a Bird's Eye View (BEV) with two additional vertical planes. Additionally, Symphonies \cite{jiang2023symphonize} highlights the significance of instance representation in SSC tasks. While SSC methods can be directly applied to 3D occupancy prediction, two primary distinctions exist: (1) SSC primarily aims to infer the occupancy of non-visible areas based on visible regions, in contrast to 3D occupancy prediction which focuses on visible areas; (2) SSC methods usually target static scenes, whereas 3D occupancy prediction methods are often designed to handle dynamic scenes.

Most existing volume-based SSC and 3D occupancy prediction methods are characterized by their dense nature, encompassing inputs and outputs that span the entire scene. Consider the BEVFormer baseline as an example: it segments the scene into uniform BEV grids, failing to distinguish between grids in varying areas. This uniformity restricts the ability of volume-based methods like BEVFormer to sample areas of interest for better performance during training. Besides, if we want to focus on a specific area in the inference stage, volume-based methods are limited to infer the entire scene and then perform post-processing, inevitably leading to increased and unnecessary costs. Moreover, as scene size and voxel resolution increase, the computational demands skyrocket exponentially. In stark contrast, our point-based model introduces much-needed flexibility by focusing on PoIs. Our method facilitates direct inference in specific areas, eliminating the need for post-processing and avoiding additional computational burdens.

A point-related SSC method is PointOcc \cite{pointocc}, a point cloud-based SSC prediction method using three complementary view planes for efficient point cloud feature modeling and an efficient 2D backbone for processing to reduce computational load, while our method focuses on the flexibility of training and inference.

\subsection{Camera-based 3D Detection}
Camera-based 3D perception tasks have gained substantial attention in recent research, largely due to the convenience and cost-effectiveness of cameras as data collection sensors. Initial efforts, such as FCOS3D \cite{fcos3d} and DETR3D \cite{wang2022detr3d}, explore the transition from 2D to 3D predictions. BEVFormer \cite{li2022bevformer} represents a significant advancement in this area, transforming images captured from vehicle-mounted cameras into a bird's eye view (BEV) representation. This technique not only enhances a vehicle's environmental understanding but also finds applications in various downstream tasks like BEVStereo \cite{bevstereo} and BEVDet \cite{bevdet} and extends to 3D occupancy prediction.

3D detection tasks are important in camera-based 3D perception and also have a high similarity to 3D occupancy prediction. The primary objective of 3D detection involves estimating the position and dimensions of objects in 3D space. DETR3D \cite{wang2022detr3d}, drawing inspiration from DETR \cite{zhu2020deformable}, innovatively combines 3D object queries with image features, incorporating camera intrinsic and extrinsic parameters. PETR \cite{liu2022petr} and its successor, PETRv2 \cite{liu2022petrv2}, further refine this approach by addressing the accuracy of reference point sampling and the inclusion of global information, enhancing 3D detection through historical data integration. Compared to object detection, 3D occupancy prediction offers a finer granularity, which is crucial for navigating irregular obstacles or overhanging objects. It is imperative to explore how insights from BEV representation and 3D object detection can inspire innovative solutions tailored to the specific requirements of 3D occupancy prediction.

\section{Preliminary}\label{subsec:pre}

\paragraph{\textbf{Problem setup.}} 
We aim to provide occupancy predictions around the ego-vehicle, given only $N$ surround-view RGB images. More specifically, we use as input current images denoted by $\mathbf{I}_i = \{ I_0, I_1, ..., I_N\}$, and use as output an occupancy prediction $\mathbf{Y}_i \in \{c_0, c_1, ..., c_M \}^{H \times W \times Z}$ defined in the coordinate of ego-vehicle, where each occupancy prediction is either empty (denoted by $c_0$) or occupied by a certain semantic class in $\{c_1, c_m, ..., c_M \}$. We assume known camera intrinsic parameters $\{K_{i}\}$ and extrinsic parameters $\{[R_i|t_i]\}$ in each frame. We assume to know whether each area is visible or not by applying a camera visibility mask.

\paragraph{\textbf{Points of Interest.}} \label{pre:poi}
In our model, we innovatively introduce the concept of Points of Interest (PoIs), which is a set of sparse points to represent the 3D scene. PoIs can flexibly represent objects or regions that require extra attention such as pedestrians or regions near ego-vehicles and can be designed as needed in both training and inference phases. We use three types of PoIs and the definitions and functions are introduced as follows:

(1) \textit{Standard Grids:} By sampling center points of grids in the inference stage and making predictions of standard 3D occupancy grids, our model makes a fair comparison with existing methods and achieves good results.

(2) \textit{Adaptively Sampling:} During the training stage, our model adaptively samples points and oversamples points around them to enhance accuracy. Recognizing that volume-based methods uniformly treat all locations, our point-based approach allows for resampling in either areas of special interest or those that are challenging to learn. This adaptively resampling strategy is also used to augment the performance of volume-based methods. Consequently, our method can function as a versatile plugin, seamlessly integrating with and enhancing existing volume-based approaches.

(3) \textit{Manually Sampling:} Our model excels in its flexibility to sample any area, specifically catering to the unique demands of various downstream tasks. Our model can make predictions of areas beyond the standard perception range by setting PoIs to areas outside the scene manually, \textit{e.g.} 200 meters away from ego-vehicle, which is a feat unattainable by traditional volume-based methods. This extension not only broadens the scope of inference but also introduces a new dimension to scene understanding.

These PoIs are the foundation of our method, offering precision and high flexibility.

\section{Method}

\begin{figure*}
    \centering
    \includegraphics[width=0.925\linewidth]{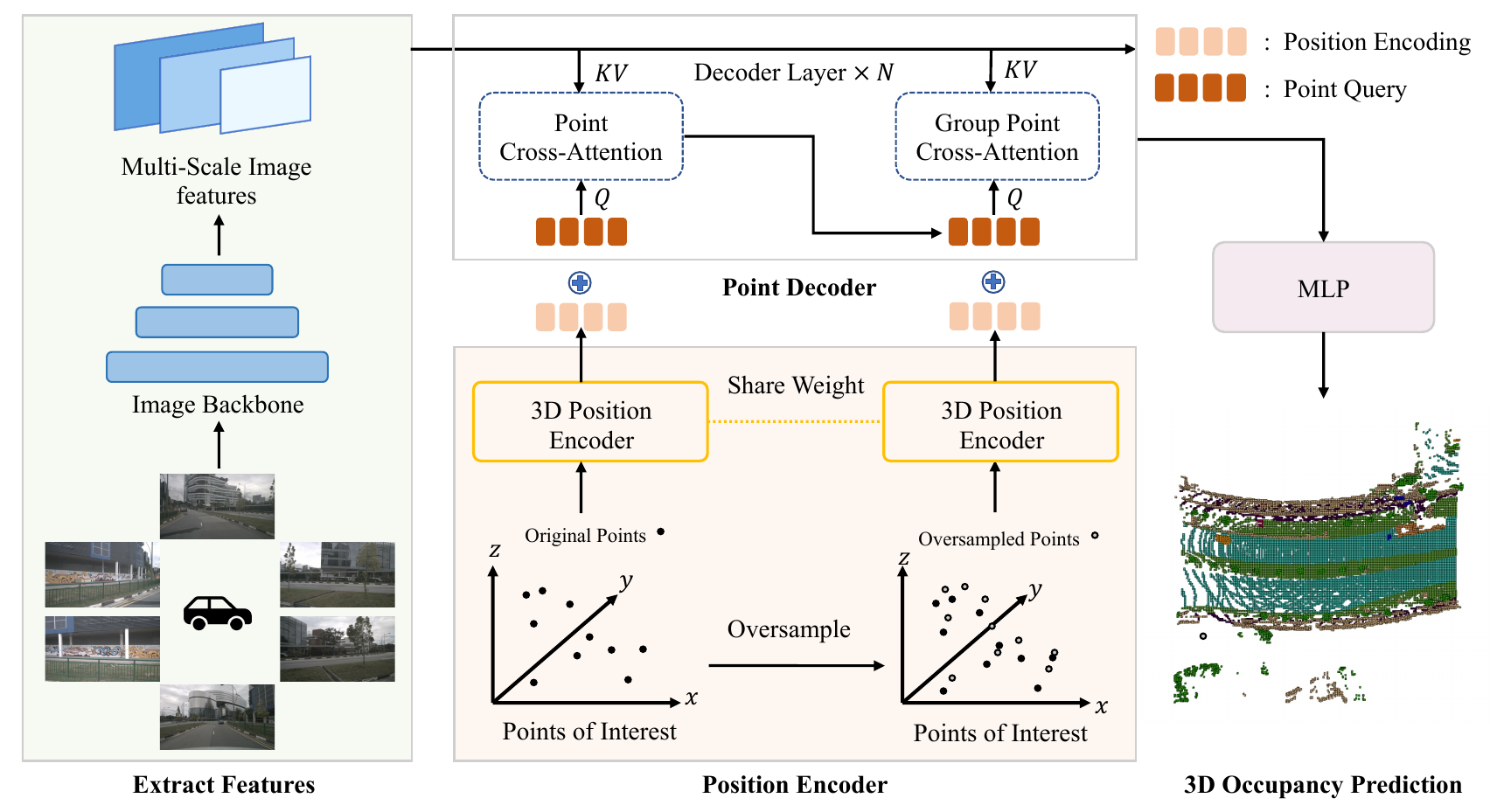}
    \caption{\textbf{Overall framework of Occupancy as Set of Points.} \name{} leverages the Transformer architecture to derive 3D point features from 2D images to make 3D occupancy predictions. Initially, we extract 2D features from multi-view images. Following this, we employ a set of 3D point queries to index these 2D features. The selection of these 3D point queries depends on the Points of Interest (PoIs).}
    \label{fig:main}
\end{figure*}

\subsection{Overall Architecture} 

Fig. \ref{fig:main} shows the overall architecture of the proposed Occupancy as Set of Points. Given the images $\mathbf{I} = \{ I_i,i = 1,2,...,N \}$ from N views, We feed these images into ResNet \cite{he2016deep} to obtain their features. 

\name{} begins with sampling a set of 3D points in the space, which are the initial PoIs. In our experiments, we use center points of grids as the initial PoIs to make a fair comparison with traditional volume-based methods and provide a reliable baseline for performance comparison and evaluation. We sample $K (K = 8000)$ points within the camera's visible region and introduce random perturbations to these points.

Then we normalize the 3D points. To these normalized coordinates, we apply sine and cosine functions as a form of positional encoding. This encoded positional information is then utilized to create query position embeddings. In the training phase, within each decoder layer, the query position corresponding to each individual query remains consistent. The 3D coordinates, along with the camera's intrinsic and extrinsic parameters, are used to map these points onto the pixel plane. This mapping process yields corresponding key and value pairs. Subsequently, we employ point cross-attention mechanisms to compute the output. Then we adaptively oversample a group of $M$ points $(M = 8000)$ whose coordinates are calculated by a linear layer and employ group point cross-attention to fuse the features of the additional sampling points.

\subsection{3D Position Encoder} 
After obtaining 3D points by applying PoIs, we first normalize coordinate points using equations as follows:
\begin{equation}
{x, y, z} = \frac{x - x_{\min}}{x_{\max} - x_{\min}},  \frac{y - y_{\min}}{y_{\max} - y_{\min}}, \frac{z - z_{\min}}{z_{\max} - z_{\min}} ,
\end{equation}
\label{eq1}
where [$x_{\min}$, $y_{\min}$, $z_{\min}$, $x_{\max}$, $y_{\max}$, $z_{\max}$] = $[-40m, -40m, -1m, 40m, 40m, 5.4m]$, are the preset boundaries of the scene. Through sine and cosine functions, we encode the normalized coordinates into high-dimensional positional information. Then we use a small MLP containing two linear layers and one ReLU layer to transform the high-dimensional positional information into learnable embeddings.

\subsection{Point Decoder} 
We used three decoder layers to build our point decoder. Each layer of the decoder contains point cross-attention (PCA) and group point cross-attention (GPCA).

The purpose of PCA and GPCA is to integrate position embedding with image features, thereby facilitating a more cohesive representation. Owing to the high efficiency of \textbf{deformable attention}~\cite{li2022bevformer}, our point decoder mechanisms employ this technique. Consequently, each query within our framework can be updated as follows:
\begin{equation}
    \texttt{DA}(\mathbf{q}, \mathbf{p}, \mathbf{F}) = \sum_{s=1}^{N_s} \mathbf{A}_{s} \mathbf{W}_s \mathbf{F^{2D}}(\mathbf{p}+\Delta \mathbf{p}_{s}),
\end{equation}
where $N_s$ represents the number of sampling offsets,  $\mathbf{A}_{s}$ represents the learnable attention weights, $\mathbf{F^{2D}}(\mathbf{p}+\Delta \mathbf{p}_{s})$ represents the image features collected at location $\mathbf{p}+\Delta \mathbf{p}_{s}$ in which $\Delta \mathbf{p}_{s}$ represents the offset apply to position  $\mathbf{p}$.

\paragraph{\textbf{Point cross-attention.}}
The point cross-attention mechanism begins by taking point queries, initially set to zero, and combines them with point position encoding derived from our 3D position encoder. This combination forms the input queries. Subsequently, these queries undergo deformable cross-attention with 2D image features. Given that not all 3D points project onto the image plane, particularly in the context of the six surround views provided by the nuScenes Dataset. Each point is likely to map onto only one or two of these views. We utilize the camera's intrinsic and extrinsic parameters to ascertain which images a given point can map to. This approach ensures that for any specific point, we only consider the features of the image(s) it maps onto, significantly reducing memory consumption. As we directly derive 3D points and generate point query $\mathbf{q}$ through our 3D Position Encoder, the projection mapping of 3D points onto 2D image features is efficiently executed.
$\mathbf{F}^{2D} = \{ \mathbf{F}^{2D}_t, \mathbf{F}^{2D}_{t-1}, ...\}$ represents the mapped 2D image features where $t$ indexes the images.
Therefore, the formula of the point cross-attention can be described as follows:
\begin{equation}
\small
    \texttt{PCA}(\mathbf{q}, \mathbf{F}^{2D} ) = \frac{1}{|\mathcal{V}_{t}|} \sum_{t \in \mathcal{V}_{t}} \texttt{DA}(\mathbf{q}, \mathcal{P}(\mathbf{p},t), \mathbf{F}_t^{2D}),
\end{equation}
where $\mathcal{V}_{t}$ represents the hit image, $t$ indexes the images, $\mathcal{P}(\mathbf{p},t)$ represents the projection function of input position $\mathbf{p}$. 

\paragraph{\textbf{Group point cross-attention.}}
The group point cross-attention mechanism is aimed at solving the lack of local context in PCA since each point independently interacts with the image features. We adaptively oversample a group of $M$ $(M = 8000)$ points around our PoIs whose coordinates are calculated by a linear layer. We use the attention obtained from PCA and the 2D image features mapped by our group of points. Therefore, the formula of the group point cross-attention can be described as follows:
\begin{equation}
\small
    \texttt{GPCA}(\mathbf{q_g}, \mathbf{F}^{2D} ) = \frac{1}{|\mathcal{V}_{t}|} \sum_{t \in \mathcal{V}_{t}} \texttt{DA}(\mathbf{q_g}, \mathcal{P}(\mathbf{p_g},t), \mathbf{F}_t^{2D}),
\end{equation}
where $\mathbf{q_g}$ represents attention calculated by PCA and $\mathbf{p_g}$ represents the input position of the oversampled group of points.

\subsection{Loss Function } 
We apply class-wise cross-entropy loss and dice loss to this task. The ground truth $\hat{\mathbf{G}}_t \in \{c_0, c_1, ..., c_M \}$ represent semantic information of a group of spatial points.  The class-wise cross-entropy loss can be computed by:
\begin{equation}\label{eq:csc}
\mathcal{L}_{ce} = - \sum_{n=1}^{N} \sum_{c=c_0}^{c_M} w_c {\hat{g}}_{n,c} log(\frac{e^{{g}_{n,c}}}{\sum_c e^{{g}_{n,c}}}) ,
\end{equation}
where $n$ is the index of points, $N$ is the number of selected points, $c$ indexes class, ${g}_{n,c}$ is the predicted logits for the $n$-th point belonging to class $c$, $w_c$ is a weight for each class according to the inverse of the class frequency as in~\cite{roldao2020lmscnet}. The dice loss can be computed by: 
\begin{equation}
\mathcal{L}_{dice}= 1 - \frac{2 \sum_{i=1}^{N} p_{i} g_{i}}{\sum_{i=1}^{N} p_{i}^2 + \sum_{i=1}^{N} g_{i}^2} ,
\end{equation}
where $p_{i}$ is the predicted probability for the point and $g_{i}$ is the ground truth binary label for the point. Our final loss is the sum of the two losses:
\begin{equation}
\mathcal{L}_{all}=\mathcal{L}_{dice} + \mathcal{L}_{ce} .
\end{equation}

\section{Experiment}

\subsection{Experimental Setup}
\paragraph{\textbf{Dataset.}} We perform our experiments on the nuScenes Dataset with annotation provided by Occ3D. Occ3D-nuScenes benchmark is interested in a volume of 80.0m to the front and back of the car, 80.0m to the left and right side, and 6.4m in height. Each sample is divided as a group of 3D voxel grids with a dimension of $[200, 200, 16]$ since each voxel has a size of $[0.4m, 0.4m, 0.4m]$. There are 18 categories of occupancy semantics, of which the 18th category is empty and does not participate in the evaluation. Occ3D-nuScenes provides ground-truth semantical voxel grids by aggregate points of the static scenes and moving objects, respectively. Furthermore, Occ3D-nuScenes utilizes ray-casting-based methods to estimate camera visibility and provides camera visibility masks.

\paragraph{\textbf{Evaluation metric.}} For this task, we use mIoU as the evaluation metric, which can be formulated as follows: 
\begin{equation}
mIoU=\frac{1}{C} \sum_{c=1}^C \frac{T P_c}{T P_c+F P_c+F N_c} .
\end{equation} 
Following the definition of 3D occupancy prediction, we only evaluate our results in visible regions.

\paragraph{\textbf{Implementation details.}} 
We input six images of the original size into ResNet101 to obtain image features. Our image backbone is pre-trained on the FCOS3D \cite{fcos3d}, which is the same backbone used in the BEVFormer baseline. Then the features will be taken by FPN to produce multi-scale feature maps. The dimension of features is $d=256$. We stacked three decoder layers, with point cross-attention and group point cross-attention in each decoder layer. All the cross-attentions are deformable, and we sample 8 points for each reference point on the pixel plane. There is a small MLP as our point-based occupancy predictor that projects the 256 feature dimension to the number of classes.
We train our model on 8 NVIDIA 3090 GPUs with $24$ epochs. We use the AdamW optimizer with a learning rate of $2\times10^{-4}$ and a weight decay of 0.01. The learning rate of the backbone is 10 times smaller. Besides, we conducted a separate experiment in which we trained our decoder with a frozen BEVFormer baseline backbone using the same setting above and this decoder will be used as a plugin module for the BEVFormer baseline.
 
\paragraph{\textbf{Baseline methods.}}
We compare Occupancy as Set of Points with existing methods replicated by Occ3D on Occ3D-nuScenes benchmark, including MonoScene \cite{cao2022monoscene}, TPVFormer \cite{huang2023triperspective}, BEVDet \cite{bevdet}, OccFormer \cite{zhang2023occformer} and BEVFormer \cite{li2022bevformer}. To make a fair comparison with the baseline, we provide results using BEVFormer by our implementation using the camera visible mask during the training.

\begin{table*}[t]
\caption{\textbf{3D Occupancy prediction performance on the Occ3D-nuScenes dataset}. * means the performance is achieved by our implementation using the camera mask during training. $^\dag$ means the performance is achieved with a frozen BEVFormer baseline backbone. } 
\resizebox{\textwidth}{!}{
\newcommand{\classfreq}[1]{{~\tiny(\semkitfreq{#1}\%)}}  

\def\mystrut{\rule{0pt}{1.5\normalbaselineskip}}
\begin{tabular}{l | c c c c c c c c c c c c c c c c c |c}

    \toprule
    Method 
    & \rotatebox{90}{others} 
    & \rotatebox{90}{barrier}
    & \rotatebox{90}{bicycle} 
    & \rotatebox{90}{bus} 
    & \rotatebox{90}{car} 
    & \rotatebox{90}{Cons. Veh} 
    & \rotatebox{90}{motorcycle} 
    & \rotatebox{90}{pedestrian} 
    & \rotatebox{90}{traffic cone} 
    & \rotatebox{90}{trailer} 
    & \rotatebox{90}{truck} 
    & \rotatebox{90}{Dri. Sur} 
    & \rotatebox{90}{other flat} 
    & \rotatebox{90}{sidewalk} 
    & \rotatebox{90}{terrain} 
    & \rotatebox{90}{manmade} 
    & \rotatebox{90}{vegetation} 
    & \rotatebox{90}{mIoU}  \\
    \midrule
    
    MonoScene~\cite{cao2022monoscene}  & 1.75 & 7.23 & 4.26 & 4.93 & 9.38 & 5.67 & 3.98 & 3.01 & 5.90 & 4.45 & 7.17 & 14.91 & 6.32 & 7.92 & 7.43 & 1.01 & 7.65 & 6.06 \\
    TPVFormer~\cite{huang2023triperspective}  & 7.22 & 38.90 & 13.67 & 40.78 & 45.90 & 17.23 & 19.99 & 18.85 & 14.30 & 26.69 & 34.17 & 55.65 & 35.47 & 37.55 & 30.70 & 19.40 & 16.78 & 27.83 \\
    BEVDet ~\cite{bevdet}  & 4.39 & 30.31 & 0.23 & 32.26 & 34.47 & 12.97 & 10.34 & 10.36 & 6.26 & 8.93 & 23.65 & 52.27 & 24.61 & 26.06 & 22.31 & 15.04 & 15.10 & 19.38 \\
    OccFormer~\cite{zhang2023occformer}  & 5.94 & 30.29 & 12.32 & 34.40 & 39.17 & 14.44 & 16.45 & 17.22 & 9.27 & 13.90 & 26.36 & 50.99 & 30.96 & 34.66 & 22.73 & 6.76 & 6.97 & 21.93 \\
    BEVFormer~\cite{li2022bevformer}  & 5.85 & 37.83 & 17.87 & 40.44 & 42.43 & 7.36 & 23.88 & 21.81 & 20.98 & 22.38 & 30.70 & 55.35 & 28.36 & 36.0 & 28.06 & 20.04 & 17.69 & 26.88 \\
    CTF-Occ~\cite{tian2023occ3d}  & 8.09 & 39.33 & 20.56 & 38.29 & 42.24 & 16.93 & 24.52 & 22.72 & 21.05 & 22.98 & 31.11 & 53.33 & 33.84 & 37.98 & 33.23 & 20.79 & 18.0 & 28.53 \\

    SurroundOcc \cite{wei2023surroundocc} &8.7	&39.2	&19.7	&41.4	&46.2	&18.7	&20.6	&26.4	&23.3	&27.0	&32.5	&78.0	&38.3	&46.6	&49.6	&36.7	&31.6 &34.4	\\
    OpenOccupancy \cite{wang2023openoccupancy}  &10.4 &45.7 &23.6 &42.4 &49.3 &14.8 &24.6 &27.7 &27.8 &27.6 &33.3 &79.2 &39.8 &47.1 &50.5 &37.7 &31.8 &36.1\\
    BEVDet-depth \cite{huang2021bevdet} &6.6	&41.2	&7.0	&42.7	&48.4	&18.4	&12.9	&22.0	&18.2	&28.2	&33.2	&80.1	&39.7	&49.1	&52.1	&39.9	&33.8 &33.7\\
    BEVDet-stereo \cite{huang2021bevdet} &8.6	&45.9	&14.3	&46.0	&51.2	&\textbf{23.8}	&18.9	&24.1	&22.3	&33.6	&37.9	&81.5	&40.5	&\textbf{52.6}	&\textbf{55.9}	&\textbf{46.9}	&\textbf{41.2} &38.0 \\
    BEVFormer*~\cite{li2022bevformer}  & 8.54 & 46.24 & 20.28 & 47.46 & 53.04 & 19.33 & 25.39 & 26.16 & 25.1 & 33.45 & 37.75 & 81.17 & 38.64 & 49.32 & 52.54 & 42.73 & 36.13 & 37.84 \\
    
    \midrule

    \textbf{OSP}  & \textbf{11.2} & 47.25 &27.06 & 47.57 & 53.66 & 23.21 & 29.37 & 29.68 & 28.41 & 32.39 & 39.94 & 79.35 & 41.36 & 50.31 & 53.23 & 40.52 & 35.39 & 39.41 \\
    \textbf{OSP$^\dag$}  & 8.87 & 46.33 & 21.32 & 47.51 & 53.14 & 19.6 & 26.12 & 26.84 & 26.68 & 33.67 & 37.94 & 81.21 & 39.13 & 49.48 & 52.76 & 42.73 & 36.12 & 38.20 \\
    \textbf{BEVFormer w/ OSP$^\dag$} & 10.95 & \textbf{49.0} & \textbf{27.68} & \textbf{50.24} & \textbf{55.99} & 22.96 & \textbf{31.02} & \textbf{30.91} & \textbf{30.25} & \textbf{35.6} & \textbf{41.23} & \textbf{82.09} & \textbf{42.59} & 51.9 & 55.1 & 44.82 & 38.17 & \textbf{41.21} \\

\bottomrule

\end{tabular}}
\scriptsize
\label{table:performance}
\end{table*}

\subsection{Performance}

\paragraph{\textbf{Evaluation strategy.}}
Our evaluation strategy for the method comprehensively utilized the three types of Points of Interest (PoIs) previously outlined. This approach entailed:

(1) Standard Grids. We conducted sampling at the centers of grids, enabling us to benchmark our method's mIoU against traditional volume-based methods. This comparison provided a direct assessment of our method's performance in a standard scenario.

\begin{figure*}
    \centering
    \includegraphics[width=0.9\textwidth]{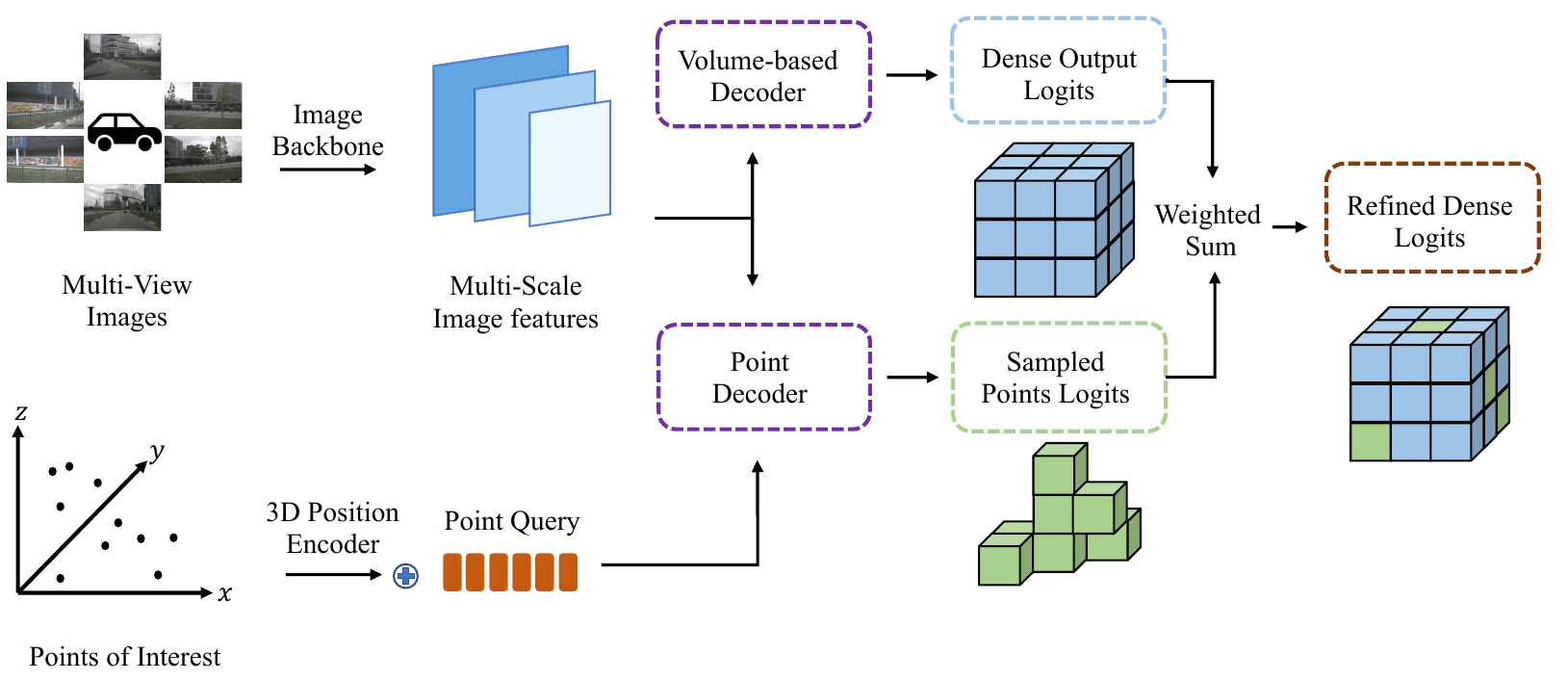}
    \caption{\textbf{Pipeline of refining volume-based methods with OSP. } Given RGB images, 2D features are extracted by the frozen image backbone of the volume-based method which in our case is the BEVFormer baseline. We use the volume-based decoder to infer the entire scene, the point decoder to infer our selected 3D points, and combine the results of both using a weighted sum method}
    \label{fig:refine}
\end{figure*}

(2) Adaptively Sampling. We adaptively oversample in our training phase to enhance our performance. Furthermore, we integrated our method as a complementary tool with volume-based approaches. By adaptively selecting PoIs that need refinement, our method enhances and augments existing volume-based techniques as shown in Fig.~\ref{fig:refine}. 

(3) Manually Sampling. We manually select points out of the standard perception range and test out the model's perceptual ability.

Through these varied and thorough evaluation techniques, we are able to comprehensively assess the performance and flexibility of our method across different scenarios and use cases

\paragraph{\textbf{Standard grids.}}
In Tab. \ref{table:performance}, we benchmarked our method against existing camera-based 3D occupancy prediction techniques on the Occ3D-nuScenes benchmark by setting PoIs to standard grids. Our point-based approach achieved a notable improvement, \textbf{obtains 1.57 mIoU performance gain} compared to our implementation of the Bevformer baseline. Notably, our method outperformed the BEVFormer baseline across almost all category metrics, demonstrating a particularly clear advantage in detecting small targets, such as the bicycle ($20.28\rightarrow27.06$), motorcycle ($25.39\rightarrow29.37$), pedestrian ($26.16\rightarrow29.37$) and traffic cone  ($25.1\rightarrow28.31$). This advantage is mainly because the direct sampling of spatial points is beneficial for feature extraction and mapping of small objects.

\paragraph{\textbf{Adaptively Sampling.}}

(1) The results shown in Tab. \ref{tab:modelarch} demonstrate that our adaptively oversampling strategy improved the results from $38.01$ to $38.48$.
(2) In the adaptive refinement experiment, we use our own implementation of the BEVFormer baseline as a representative of volume-based methods and enhance it with our method using different scales of PoIs by adaptively selecting locations with low confidence. Our model is trained with the frozen BEVFormer baseline backbone. The experimental results, as shown in Tab. \ref{tab:refine}, confirm that our plugin model significantly enhances BEVformer's performance. Notably, the BEVFormer w/ OSP results surpass those obtained from the two models operating independently. Furthermore, as we increase the size of the refined scene, the mIoU scores of BEVFormer w/ OSP also exhibit marked improvements, indicating that our refinement approach is very effective across various areas.

\paragraph{\textbf{Manually Sampling.}}

The Occ3D-nuScenes dataset offers annotations across an $[80m \times 80m]$ range. We manually set the points and annotations within a smaller $[60m \times 60m]$ area and set PoIs to points out of it. To comprehensively evaluate our method, we conducted assessments across three distinct ranges: the standard $[60m \times 60m]$ range, the full dataset range of $[80m \times 80m]$, and an intermediate-range spanning from $60m$ to $80m$. Results shown in Tab. \ref{tab:outer} demonstrate the capability of our method to make predictions at manually selected locations which is outside the predefined range in this case.

\begin{table}[h]
    \centering
    \caption{\textbf{Predict regions beyond the scene.}  The model was trained on scenes within a \(60m \times 60m\) range, and evaluations were conducted on the training scenes, the entire size of the scene, and the distance from 60m to 80m.}
    \begin{tabularx}{0.9\columnwidth}{YYYY}
    \toprule
    {Method} & {Scene}      & {Range} & {mIoU}  \\
    \midrule
    {OSP} & Standard          &  {$0m\sim60m$}    & 42.29  \\
    {OSP} & Large              & {$0m\sim80m$}  & 29.00  \\
    {OSP} & Beyond     &  {$60m\sim80m$}   & 23.00  \\
    \bottomrule
    \end{tabularx}
    \label{tab:outer}
\end{table}

\begin{table}[H]
    \centering
    \caption{\textbf{Refine BEVFormer with our decoder. }$^\dag$ means our model is trained with a frozen BEVFormer baseline backbone. Refine scale refers to an area defined by the length multiplied by the width, centered around the self-vehicle. The entire scene spans a scale of \(80m \times 80m\). As the refine scale increases, the mIoU also gradually increases.}\small
    \begin{tabularx}{0.9\columnwidth}{YYY}
    \toprule

        Method                      & {Refine Scale} & {mIoU} \\
        \midrule 
        BEVFormer                   & -                     & 37.84 \\
        OSP$^\dag$                      & -                     & 38.20 \\
        \midrule 
        {BEVFormer w/ OSP}                    & {\(20m \times 20m\)} & 38.20 \\ 
        {BEVFormer w/ OSP}                    & {\(40m \times 40m\)} & 39.35 \\ 
        {BEVFormer w/ OSP}                    & {\(50m \times 50m\)} & 40.32 \\ 
        {BEVFormer w/ OSP}                    & {\(80m \times 80m\)} & \textbf{41.21} \\ 
    
    \bottomrule
    \end{tabularx}
    
    \label{tab:refine}
\end{table}

\subsection{Ablation studies}

\paragraph{\textbf{Model architecture.}}
We conducted ablation experiments on the model structure, particularly exploring variations in the number of transformer layers and the adaptively oversampling strategy. The results of these experiments are presented in Tab. \ref{tab:modelarch}. Even without an oversampling strategy and with just a 3-layer transformer, our method's baseline already surpasses the BEVFormer baseline, advancing from $37.84$ to $38.01$ in mIoU. The oversampling strategy further amplifies this improvement by enhancing the connections between spatial points, culminating in an increased mIoU of $38.48$.

\begin{table}[h]
    \centering
    \caption{\textbf{Model architecture.} To make a fair comparison with the baseline, we set the number of layers to three. The oversampling strategy brings 0.47 mIoU.}
    \begin{tabularx}{0.9\columnwidth}{YYY}
    \toprule
        {Layer Num}  & {Oversample}                & {mIoU} \\
        \midrule 
        1                    & -                      &  36.76 \\
        3                    & -                      &  38.01 \\
        3                    & \checkmark             &  38.48 \\

    \bottomrule
    \end{tabularx}
    \label{tab:modelarch}
\end{table}

\paragraph{\textbf{2D image features.}}
The performance of our method is heavily reliant on the quality of image features. To elucidate this dependency, we conducted ablation studies focusing on the output from the model's 'neck' component. These experiments demonstrated the substantial impact of using a multi-scale output from the Feature Pyramid Network (FPN). This architectural choice notably boosts the model's performance, as evidenced by the increase in metric scores from $38.67$ to $39.41$ as shown in Tab. \ref{tab:imgfeat}.

\begin{table}[h]
    \centering
    \caption{\textbf{2D image features.} Increasing the multi-scale feature maps of images from 2 to 4 brings 0.74 mIoU.}
    \small
    \begin{tabularx}{0.9\columnwidth}{YY}
    \toprule
    
        {Multi-scale Images Features}      & {mIoU} \\
        \midrule 
         2                     &  38.67\\
         4                     &  39.41 \\

    \bottomrule
    \end{tabularx}
    \label{tab:imgfeat}
\end{table}

\paragraph{\textbf{Grid center sampling method.}}
Our ablation experiments also investigated the technique for the grid center sampling method during training. While initially using grid center points as coordinates, we introduced a variation by adding random perturbations. The results, as detailed in Tab. \ref{tab:disturb}, indicate the introduction of this disturbance to the grid center points elevates the mIoU to $38.67$.

\begin{table}[h]
    \centering
    \caption{\textbf{Point Perturbation.} We randomly applied perturbations of \(0.1m\) to the coordinates of each point in three directions. This strategy brings 0.19 mIoU.}
    
    \small
    \begin{tabularx}{0.9\columnwidth}{YY}
    \toprule
    
        {Point Perturbation}  & {mIoU} \\
        \midrule 
        -                &  38.48 \\
        \checkmark       &  38.67 \\
    
    \bottomrule
    \end{tabularx}
    \label{tab:disturb}
\end{table}

\paragraph{\textbf{Loss design.}}
The task of 3D occupancy prediction bears similarities to segmentation tasks. In the context of the Occ3D-nuScenes dataset, there is a notable issue of class imbalance. To address this challenge, we opted to incorporate dice loss into our optimization strategy. The effectiveness of this choice is evident in the results presented in Tab. \ref{tab:dice}, which demonstrate that the integration of dice loss benefits our method.

\begin{table}[h]
    \caption{\textbf{Loss design.} Incorporating dice loss for optimization brings 0.53 mIoU.}
    \centering
    \small
    \begin{tabularx}{0.9\columnwidth}{YY}
    \toprule
    
        {Dice Loss}  & {mIoU} \\
        \midrule 
        -                &  38.88 \\
        \checkmark       &  39.41 \\
    
    \bottomrule
    \end{tabularx}
    \label{tab:dice}
    \vspace{-10pt}
\end{table}

\paragraph{\textbf{Adaptive point sampling.}}
During the inference process, we apply adaptive sampling (adaptively select points with high uncertainty) in BEVFormer w/ OSP and \name{}. Adaptive sampling can reduce computational burden while maintaining good performance as shown in Tab.~\ref{tab:adaptive}. BEVFormer w/ OSP using adaptive sampling means we refine BEVFormer by adaptively selecting points with high uncertainty (around \(20\%\)). \name{} using adaptive sampling means we only forward points with low confidence to the next decoder layer, while directly outputting the results for points with high confidence in the decoder. Points of high uncertainty are defined as those with a confidence score less than a threshold of \(0.9\) after softmax.

\begin{table}[h]
    \centering
    \caption{\textbf{Adaptability of \name{}.} `rel' denotes relative.}
    \small
    \begin{tabularx}{0.9\columnwidth}{lYYY}
    
    \toprule
        {Method}                   &{Adaptive Sampling}     & {Rel. Computation}   &  {mIoU} \\
        \midrule 
        BEVFormer                     & -               & -                 & 37.84 \\
        {BEVFormer w/ OSP}            &  -               & 1.0$\times$                 & 39.35 \\      
        {BEVFormer w/ OSP}            &\checkmark      & 0.8$\times$   & 39.22 \\
        \midrule 
        OSP                           & -               & 1.0$\times$                 & 39.41 \\
        OSP                           & \checkmark      & 0.93$\times$                & 39.08 \\
    \bottomrule
    \end{tabularx}
    \label{tab:adaptive}
\end{table}

\begin{figure}[H]
    \centering
    \includegraphics[width=0.725\textwidth]{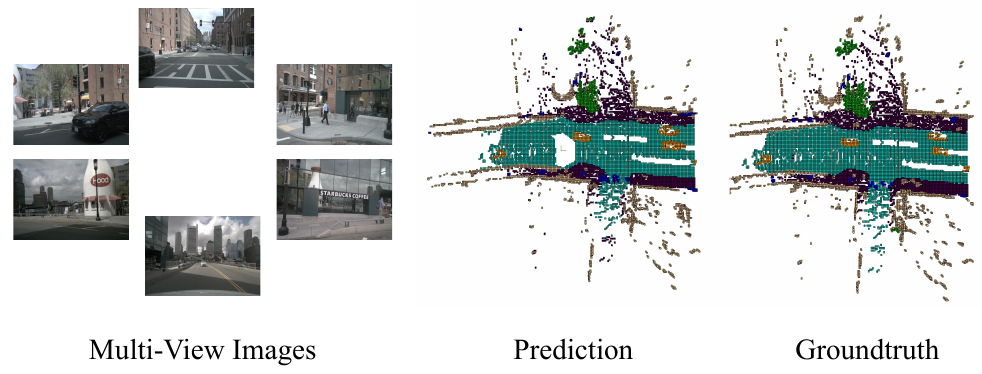}
    \caption{\textbf{Visualization of our results. } Our visualization results have voids compared to the ground truth ego-vehicle position, which is due to the surrounding view having gaps near the ego-vehicle.}
    \label{fig:vis}
\end{figure}

\section{Conclusion}
In this work, we present a novel perspective on 3D scene representation, viewing it through a set of points. We introduce the innovative concept of \textit{Points of Interest} (PoIs), which significantly advances the flexibility in scene representation. Building upon the foundation laid by PoIs, we develop a highly adaptable, point-based 3D occupancy prediction framework, named \name{}. We validate the strong performance and flexibility of \name{} on the Occ3D-nuScenes benchmark. Our work not only contributes to the field of 3D occupancy prediction but also paves the way for more dynamic and adaptable methods in 3D scene analysis.

\bibliographystyle{splncs04}
\bibliography{eccv}

\begin{thebibliography}{10}
\providecommand{\url}[1]{\texttt{#1}}
\providecommand{\urlprefix}{URL }
\providecommand{\doi}[1]{https://doi.org/#1}

\bibitem{behley2019semantickitti}
Behley, J., Garbade, M., Milioto, A., Quenzel, J., Behnke, S., Stachniss, C., Gall, J.: Semantickitti: A dataset for semantic scene understanding of lidar sequences. In: Proceedings of the IEEE/CVF International Conference on Computer Vision. pp. 9297--9307 (2019)

\bibitem{caesar2020nuscenes}
Caesar, H., Bankiti, V., Lang, A.H., Vora, S., Liong, V.E., Xu, Q., Krishnan, A., Pan, Y., Baldan, G., Beijbom, O.: nuscenes: A multimodal dataset for autonomous driving. In: Proceedings of the IEEE/CVF conference on computer vision and pattern recognition. pp. 11621--11631 (2020)

\bibitem{cai2021semantic}
Cai, Y., Chen, X., Zhang, C., Lin, K.Y., Wang, X., Li, H.: Semantic scene completion via integrating instances and scene in-the-loop. In: Proceedings of the IEEE/CVF Conference on Computer Vision and Pattern Recognition. pp. 324--333 (2021)

\bibitem{cao2022monoscene}
Cao, A.Q., de~Charette, R.: Monoscene: Monocular 3d semantic scene completion. In: Proceedings of the IEEE/CVF Conference on Computer Vision and Pattern Recognition. pp. 3991--4001 (2022)

\bibitem{chen20203d}
Chen, X., Lin, K.Y., Qian, C., Zeng, G., Li, H.: 3d sketch-aware semantic scene completion via semi-supervised structure prior. In: Proceedings of the IEEE/CVF Conference on Computer Vision and Pattern Recognition. pp. 4193--4202 (2020)

\bibitem{he2016deep}
He, K., Zhang, X., Ren, S., Sun, J.: Deep residual learning for image recognition. In: Proceedings of the IEEE conference on computer vision and pattern recognition. pp. 770--778 (2016)

\bibitem{bevdet}
Huang, J., Huang, G., Zhu, Z., Du, D.: Bevdet: High-performance multi-camera 3d object detection in bird-eye-view. arXiv preprint arXiv:2112.11790  (2021)

\bibitem{huang2021bevdet}
Huang, J., Huang, G., Zhu, Z., Du, D.: Bevdet: High-performance multi-camera 3d object detection in bird-eye-view. CoRR  (2021)

\bibitem{huang2023triperspective}
Huang, Y., Zheng, W., Zhang, Y., Zhou, J., Lu, J.: Tri-perspective view for vision-based 3d semantic occupancy prediction. In: {CVPR}. pp. 9223--9232 (2023)

\bibitem{jiang2023symphonize}
Jiang, H., Cheng, T., Gao, N., Zhang, H., Liu, W., Wang, X.: Symphonize 3d semantic scene completion with contextual instance queries. CoRR  (2023)

\bibitem{li2020anisotropic}
Li, J., Han, K., Wang, P., Liu, Y., Yuan, X.: Anisotropic convolutional networks for 3d semantic scene completion. In: Proceedings of the IEEE/CVF Conference on Computer Vision and Pattern Recognition. pp. 3351--3359 (2020)

\bibitem{li2019rgbd}
Li, J., Liu, Y., Gong, D., Shi, Q., Yuan, X., Zhao, C., Reid, I.D.: {RGBD} based dimensional decomposition residual network for 3d semantic scene completion. In: {CVPR}. pp. 7693--7702 (2019)

\bibitem{li2019depth}
Li, J., Liu, Y., Yuan, X., Zhao, C., Siegwart, R., Reid, I., Cadena, C.: Depth based semantic scene completion with position importance aware loss. IEEE Robotics and Automation Letters  \textbf{5}(1),  219--226 (2019)

\bibitem{bevstereo}
Li, Y., Bao, H., Ge, Z., Yang, J., Sun, J., Li, Z.: Bevstereo: Enhancing depth estimation in multi-view 3d object detection with temporal stereo. In: {AAAI}. pp. 1486--1494 (2023)

\bibitem{li2022bevformer}
Li, Z., Wang, W., Li, H., Xie, E., Sima, C., Lu, T., Yu, Q., Dai, J.: Bevformer: Learning bird's-eye-view representation from multi-camera images via spatiotemporal transformers. arXiv preprint arXiv:2203.17270  (2022)

\bibitem{liu2018see}
Liu, S., Hu, Y., Zeng, Y., Tang, Q., Jin, B., Han, Y., Li, X.: See and think: Disentangling semantic scene completion. In: Advances in Neural Information Processing Systems. vol.~31 (2018)

\bibitem{liu2022petr}
Liu, Y., Wang, T., Zhang, X., Sun, J.: Petr: Position embedding transformation for multi-view 3d object detection. arXiv preprint arXiv:2203.05625  (2022)

\bibitem{liu2022petrv2}
Liu, Y., Yan, J., Jia, F., Li, S., Gao, Q., Wang, T., Zhang, X., Sun, J.: Petrv2: A unified framework for 3d perception from multi-camera images. arXiv preprint arXiv:2206.01256  (2022)

\bibitem{miao2023occdepth}
Miao, R., Liu, W., Chen, M., Gong, Z., Xu, W., Hu, C., Zhou, S.: Occdepth: A depth-aware method for 3d semantic scene completion (2023)

\bibitem{moravec1985high}
Moravec, H., Elfes, A.: High resolution maps from wide angle sonar. In: Proceedings. 1985 IEEE international conference on robotics and automation. vol.~2, pp. 116--121. IEEE (1985)

\bibitem{roldao2020lmscnet}
Roldao, L., de~Charette, R., Verroust-Blondet, A.: Lmscnet: Lightweight multiscale 3d semantic completion. In: 2020 International Conference on 3D Vision (3DV). pp. 111--119. IEEE (2020)

\bibitem{song2017semantic}
Song, S., Yu, F., Zeng, A., Chang, A.X., Savva, M., Funkhouser, T.: Semantic scene completion from a single depth image. In: Proceedings of the IEEE conference on computer vision and pattern recognition. pp. 1746--1754 (2017)

\bibitem{sun2020scalability}
Sun, P., Kretzschmar, H., Dotiwalla, X., Chouard, A., Patnaik, V., Tsui, P., Guo, J., Zhou, Y., Chai, Y., Caine, B., et~al.: Scalability in perception for autonomous driving: Waymo open dataset. In: Proceedings of the IEEE/CVF conference on computer vision and pattern recognition. pp. 2446--2454 (2020)

\bibitem{thrun2002probabilistic}
Thrun, S.: Probabilistic robotics. Communications of the ACM  \textbf{45}(3),  52--57 (2002)

\bibitem{tian2023occ3d}
Tian, X., Jiang, T., Yun, L., Mao, Y., Yang, H., Wang, Y., Wang, Y., Zhao, H.: Occ3d: A large-scale 3d occupancy prediction benchmark for autonomous driving (2023)

\bibitem{fcos3d}
Wang, T., Zhu, X., Pang, J., Lin, D.: Fcos3d: Fully convolutional one-stage monocular 3d object detection. arXiv preprint arXiv:2104.10956  (2021)

\bibitem{wang2023openoccupancy}
Wang, X., Zhu, Z., Xu, W., Zhang, Y., Wei, Y., Chi, X., Ye, Y., Du, D., Lu, J., Wang, X.: Openoccupancy: A large scale benchmark for surrounding semantic occupancy perception. arXiv preprint arXiv:2303.03991  (2023)

\bibitem{wang2022detr3d}
Wang, Y., Guizilini, V.C., Zhang, T., Wang, Y., Zhao, H., Solomon, J.: Detr3d: 3d object detection from multi-view images via 3d-to-2d queries. In: Conference on Robot Learning. pp. 180--191. PMLR (2022)

\bibitem{wei2023surroundocc}
Wei, Y., Zhao, L., Zheng, W., Zhu, Z., Zhou, J., Lu, J.: Surroundocc: Multi-camera 3d occupancy prediction for autonomous driving. arXiv preprint arXiv:2303.09551  (2023)

\bibitem{zhang2018efficient}
Zhang, J., Zhao, H., Yao, A., Chen, Y., Zhang, L., Liao, H.: Efficient semantic scene completion network with spatial group convolution. In: Proceedings of the European Conference on Computer Vision (ECCV). pp. 733--749 (2018)

\bibitem{zhang2019cascaded}
Zhang, P., Liu, W., Lei, Y., Lu, H., Yang, X.: Cascaded context pyramid for full-resolution 3d semantic scene completion. In: Proceedings of the IEEE/CVF International Conference on Computer Vision. pp. 7801--7810 (2019)

\bibitem{zhang2023occformer}
Zhang, Y., Zhu, Z., Du, D.: Occformer: Dual-path transformer for vision-based 3d semantic occupancy prediction. arXiv preprint arXiv:2304.05316  (2023)

\bibitem{zhu2020deformable}
Zhu, X., Su, W., Lu, L., Li, B., Wang, X., Dai, J.: Deformable detr: Deformable transformers for end-to-end object detection. In: International Conference on Learning Representations (2020)

\bibitem{pointocc}
Zuo, S., Zheng, W., Huang, Y., Zhou, J., Lu, J.: Pointocc: Cylindrical tri-perspective view for point-based 3d semantic occupancy prediction  (2023)

\end{thebibliography}

\end{document}